\pdfoutput=1

\documentclass{article}
\usepackage{colm2024_conference}

\usepackage{url}
\usepackage{booktabs}
\usepackage{hyperref}
\usepackage{graphicx}
\usepackage{microtype}
\usepackage{multicol}
\usepackage{xcolor}   
\usepackage{multirow}
\usepackage{lipsum}
\usepackage{enumitem}
\usepackage{pifont}
\usepackage{colortbl}
\usepackage{tabularx}
\usepackage{caption} 
\usepackage{float}
\usepackage{amsmath,amsfonts,amssymb,bbm}

\newcommand*\samethanks[1][\value{footnote}]{\footnotemark[#1]}
\newcommand{\squad}{\quad\ }

\title{Soft Adaptive Policy Optimization}

\author{
\textbf{Chang Gao\thanks{Corresponding authors.} \squad Chujie Zheng \squad Xiong-Hui Chen \squad Kai Dang \squad Shixuan Liu \\ Bowen Yu\samethanks{}  \squad An Yang\samethanks{} \squad Shuai Bai \squad Jingren Zhou \squad Junyang Lin } \\
\vspace{1.5mm}
Qwen Team, Alibaba Inc.
}

\begin{document}
\maketitle

\begin{abstract}
Reinforcement learning (RL) plays an increasingly important role in enhancing the reasoning capabilities of large language models (LLMs), yet stable and performant policy optimization remains challenging. Token-level importance ratios often exhibit high variance—a phenomenon exacerbated in Mixture-of-Experts models—leading to unstable updates. Existing group-based policy optimization methods, such as GSPO and GRPO, alleviate this problem via hard clipping, making it difficult to maintain both stability and effective learning. We propose Soft Adaptive Policy Optimization (SAPO), which replaces hard clipping with a smooth, temperature-controlled gate that adaptively attenuates off-policy updates while preserving useful learning signals. Compared with GSPO and GRPO, SAPO is both sequence-coherent and token-adaptive. Like GSPO, SAPO maintains sequence-level coherence, but its soft gating forms a continuous trust region that avoids the brittle hard clipping band used in GSPO. When a sequence contains a few highly off-policy tokens, GSPO suppresses all gradients for that sequence, whereas SAPO selectively down‑weights only the offending tokens and preserves the learning signal from the near-on‑policy ones, improving sample efficiency. Relative to GRPO, SAPO replaces hard token‑level clipping with smooth, temperature-controlled scaling, enabling more informative and stable updates. Empirical results on mathematical reasoning benchmarks indicate that SAPO exhibits improved training stability and higher Pass@1 performance under comparable training budgets. Moreover, we employ SAPO to train the Qwen3-VL model series, demonstrating that SAPO yields consistent performance gains across diverse tasks and different model sizes. Overall, SAPO provides a more reliable, scalable, and effective optimization strategy for RL training of LLMs.

\end{abstract}

\section{Introduction}

Reinforcement learning (RL) has become a key driver of recent advances in large language models (LLMs), enabling deeper and longer reasoning for challenging tasks in mathematics, programming, and multimodal understanding \citep{o1,dpsk-r1,qwen3}. Among RL methods, group-based policy optimization has emerged as a practical recipe: multiple responses are sampled per query, sequence-level rewards are normalized within the group, and policy updates are weighted by importance ratios between the current and behavior policies \citep{grpo,gspo}. 

A central challenge in this setting is the high variance of token-level importance ratios, especially in Mixture-of-Experts (MoE) models where routing heterogeneity and long responses can amplify deviations across tokens. Such variance increases the likelihood of unstable updates. Hard clipping, as used in GRPO \citep{grpo}, constrains large deviations by zeroing gradients outside a fixed band. While effective at curbing excessive steps, hard clipping makes it difficult to strike a favorable trade-off: overly tight clipping limits the number of valid samples for gradient computation, while looser clipping introduces noisy gradients coming from off-policy samples.

To address the brittleness of hard clipping in group-based policy optimization, we propose \textbf{Soft Adaptive Policy Optimization (SAPO)}, a smooth and adaptive policy-gradient method that replaces hard clipping with a temperature-controlled soft gate, as shown in Figure~\ref{fig:sapo}. SAPO weights token-level updates by a bounded, sigmoid-shaped function of the importance ratio, centered at the on-policy point. This implements a continuous trust region: near on-policy, gradients are preserved to encourage useful updates and exploration; as the ratio deviates, gradients are attenuated smoothly rather than truncated, maintaining the learning signal for moderate deviations while reducing optimization noise. To further enhance robustness in large vocabularies, SAPO employs asymmetric temperatures for positive and negative tokens to make gradients on negative tokens decay more rapidly, reflecting their distinct stability profiles: negative updates tend to increase the logits of many inappropriate tokens and are therefore more prone to introduce instability than positive updates.

Conceptually, SAPO is designed to be sequence-coherent and token-adaptive. Under mild and empirically common conditions—small on-policy steps and low dispersion of token log-ratios within a sequence—the average token gate concentrates to a smooth sequence-level gate, aligning optimization with sequence-level rewards in the spirit of sequence-based methods such as GSPO \citep{gspo}. When these conditions are violated due to heterogeneous or outlier tokens, SAPO selectively down-weights only the offending tokens while retaining informative gradients from near on-policy tokens within the same sequence. This selective attenuation mitigates the signal loss associated with hard clipping, improving sampling efficiency while maintaining stable updates. 

Empirically, SAPO provides improved stability and task performance compared with GSPO and GRPO. While all methods may ultimately exhibit signs of instability, SAPO sustains coherent learning for a longer duration and reaches higher Pass@1 accuracy before divergence. This stems from SAPO’s ability to preserve informative gradients beyond the hard-clip threshold while selectively suppressing high-variance token updates. Furthermore, our temperature ablations further reveal that the asymmetric design—using a larger temperature for negative-token updates—is critical: it dampens high-variance negative gradients and significantly reduces the likelihood of early collapse. Beyond controlled settings, SAPO also proves effective in practical training of Qwen3-VL models across a broad mixture of text and multimodal tasks and across different model scales and architectures.
Together, these results demonstrate that SAPO’s smooth gating and asymmetric temperature control enable a more reliable and productive RL training of large language models.

\section{Preliminaries}

\paragraph{Notation.}
We model an autoregressive language model parameterized by $\theta$ as a stochastic policy $\pi_\theta$ over token sequences. Let $q$ denote a query and $\mathcal{D}$ the query set. For a response $y$ to $q$, its likelihood under $\pi_\theta$ factorizes as
$\pi_\theta(y \mid q) \;=\; \prod_{t=1}^{|y|} \pi_\theta\!\left(y_t \mid q, y_{<t}\right)$,
where $|y|$ is the number of tokens in $y$.

\paragraph{Group Relative Policy Optimization (GRPO).}
For each query $q \sim \mathcal{D}$, GRPO~\citep{grpo} samples a group of $G$ responses $\{y_1,\ldots,y_G\}$ from the behavior policy $\pi_{\theta_\text{old}}$, computes their rewards $\{R_1,\ldots,R_G\}$, and maximizes the following token-level objective:
\begin{align}
\label{equ:grpo}
\mathcal{J}_\text{GRPO}(\theta) = \mathbb{E}_{ q \sim \mathcal{D},\, \{y_i\}_{i=1}^G \sim \pi_{\theta_\text{old}}( \cdot | q) }
\left[ \frac{1}{G} \sum_{i=1}^{G} \frac{1}{|y_i|} \sum_{t=1}^{|y_i|} 
\min \left( r_{i,t}(\theta) \widehat{A}_{i,t},  \, \mathrm{clip} \left( r_{i,t}(\theta), 1 - {\varepsilon}, 1 + {\varepsilon}\right) \widehat{A}_{i,t} \right)
\right],
\end{align}
where 

\begin{align}
    r_{i,t}(\theta)=\frac{ \pi_{\theta} (y_{i,t} | q, y_{i,<t}) }{ \pi_{\theta_\text{old}} (y_{i,t} | q,y_{i,<t})},\squad
    \widehat{A}_{i,t} = \widehat{A}_{i} = \frac{ R_i - \text{mean}(\{R_j\}_{j=1}^G) }{ \mathrm{std} \left( \{R_j\}_{j=1}^G \right) },
\label{eq:ratio}
\end{align}

$\varepsilon>0$ is the clipping range, $G$ is the number of responses in a group, and $\widehat{A}_{i,t}$ is the group-normalized advantage (shared across tokens within a response).

\paragraph{Group Sequence Policy Optimization (GSPO)}

GSPO \citep{gspo} employs the following sequence-level optimization objective:
\begin{align}
\mathcal{J}_\text{GSPO} (\theta) =
\mathbb{E}_{ q \sim \mathcal{D},\, \{y_i\}_{i=1}^G \sim \pi_{\theta_\text{old}}( \cdot | q) }
\left[ 
\frac{1}{G} \sum_{i=1}^{G}
\min \left( s_{i}(\theta)  \widehat{A}_{i},  \, \mathrm{clip} \left( s_{i}(\theta), 1 - {\varepsilon}, 1 + {\varepsilon} \right) \widehat{A}_{i} \right) 
\right],
\label{equ:gspo}
\end{align}
where 
\begin{align}
s_{i}(\theta) = \left( \frac{ \pi_{\theta} (y_i | q) }{ \pi_{\theta_\text{old}} (y_i | q)} \right)^{\frac{1}{|y_i|}}
=
\exp \left( \frac{1}{|y_i|} \sum_{t=1}^{|y_i|} \log \frac{ \pi_{\theta} (y_{i,t} | q, y_{i,<t}) }{ \pi_{\theta_\text{old}} (y_{i,t} | q,y_{i,<t})} \right), \squad
\widehat{A}_{i} = \frac{ R_i - \text{mean}(\{R_j\}_{j=1}^G) }{ \mathrm{std} \left( \{R_j\}_{j=1}^G \right) }
\end{align}
GSPO applies clipping at the sequence level rather than per token. The length normalization in $s_i(\theta)$ reduces variance and places it on a consistent numerical scale across responses.

\section{Soft Adaptive Policy Optimization}

\begin{figure*}[t]
\centering
\includegraphics[width= 1\linewidth]{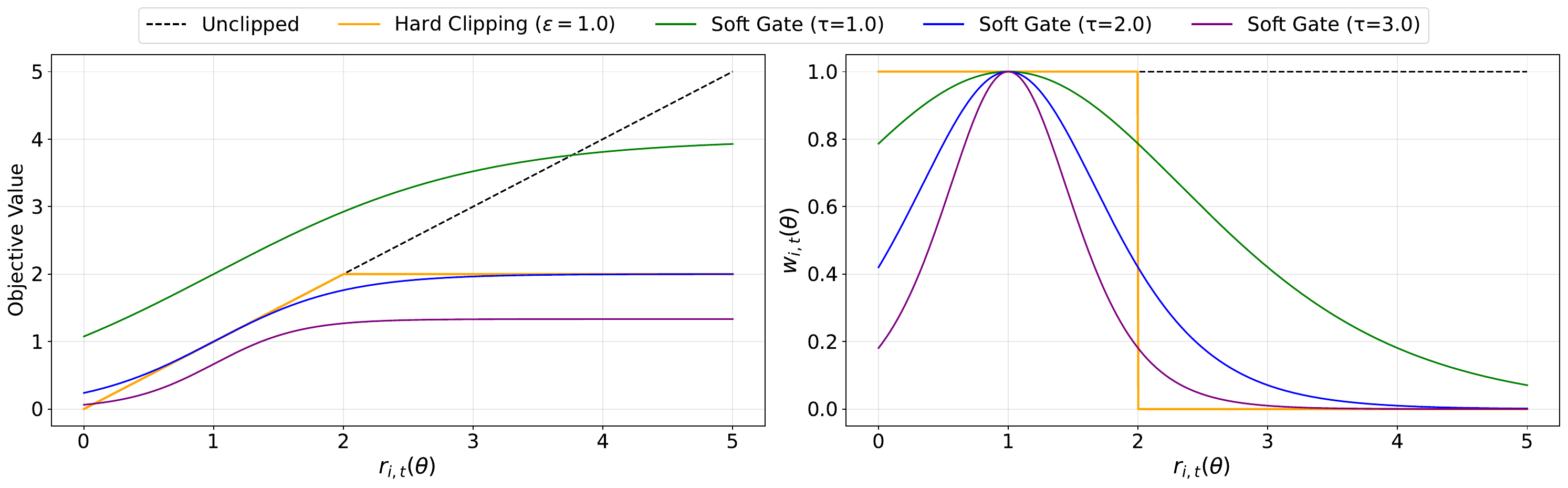}
   \caption{Comparison of policy-update objectives under positive advantage. The left panel shows the surrogate objective value; the right panel shows the corresponding gradient weight $w_{i,t}(\theta)$ as a function of the policy ratio $r_{i,t}(\theta)$.}
\label{fig:sapo}
\end{figure*}

Soft Adaptive Policy Optimization (SAPO) is a smooth and adaptive policy-gradient method for RL fine-tuning, which replaces hard clipping with a temperature-controlled soft gate. Smooth gating functions have been explored in traditional RL settings~\citep{chen2023sufficiency}. In SAPO, we incorporate this idea into the group‑based RL paradigm for LLMs and extend it with two additional components that are essential for large‑scale language model training:
(1) a token‑level soft trust region that naturally yields sequence‑level coherence, and
(2) an asymmetric temperature design motivated by the distinct behaviors of positive and negative token updates.
Specifically, SAPO maximizes the following objective:
\begin{equation}
\begin{aligned}
\mathcal{J}(\theta) = \mathbb{E}_{q\sim \mathcal{D}, \{y_i\}_{i=1}^G\sim \pi_{\theta_\text{old}}(\cdot\mid q)}
\Bigg[\frac{1}{G} \sum_{i=1}^{G} \frac{1}{|y_i|} \sum_{t=1}^{|y_i|} 
 f_{i,t}(r_{i,t}(\theta))
 \widehat{A}_{i,t}
 \Bigg],
\label{eq:sapo}
\end{aligned}
\end{equation}
where
\begin{equation}
    f_{i,t}(x) = \sigma\left(\tau_{i,t}\,(x-1)\right) \cdot \frac{4}{\tau_{i,t}}, \quad
    \tau_{i,t} = 
    \begin{cases} 
    \tau_{\text{pos}}, & \text{if } \widehat{A}_{i,t} > 0\\
    \tau_{\text{neg}}, & \text{otherwise} 
    \end{cases},
\label{eq:tau}
\end{equation}
$\widehat{A}_{i,t}$ and $r_{i,t}(\theta)$ are computed as in Equation \eqref{eq:ratio}, $\tau_{\text{pos}}$ and $\tau_{\text{neg}}$ are the temperatures for positive and negative tokens, respectively, and $\sigma(x)=1/(1+e^{-x})$ is the sigmoid function. 

Differentiating \eqref{eq:sapo} yields a weighted log-policy gradient:
\begin{equation}
\begin{aligned}
\nabla_{\theta}\mathcal{J}(\theta) & = \mathbb{E}_{q\sim \mathcal{D}, \{y_i\}_{i=1}^G\sim \pi_{\theta_\text{old}}(\cdot\mid q)}
\Bigg[\frac{1}{G} \sum_{i=1}^{G} \frac{1}{|y_i|} \sum_{t=1}^{|y_i|} 
w_{i,t}(\theta)\,r_{i,t}(\theta)\,\nabla_\theta \log \pi_\theta(y_{i,t}\mid q,y_{i,<t})
 \widehat{A}_{i,t}
 \Bigg] 
\label{eq:softgrpo_gradient}
\end{aligned}
\end{equation}
where
\begin{equation}
    w_{i,t}(\theta) = 4\,p_{i,t}(\theta)\,\bigl(1-p_{i,t}(\theta)\bigr), \quad p_{i,t}(\theta)=\sigma\left(\tau_{i,t}\,(r_{i,t}(\theta)-1)\right),
\label{eq:weight}
\end{equation}
which peaks at $r_{i,t}(\theta)=1$ with a value of $1$ and decays smoothly and approximately exponentially as $r_{i,t}(\theta)$ deviates from $1$, implementing a soft trust region and preventing both gradient vanishing and excessively large updates, as shown in Figure \ref{fig:sapo}. Notably, at $r_{i,t}(\theta)=1$, the soft-gated gradient equals that of the unclipped objective $r_{i,t}(\theta)\widehat{A}_{i,t}$, regardless of $\tau_{i,t}$, preserving on-policy behavior. This also explains the presence of the $4 / \tau_{i,t}$ factor in $f_{i,t}$.

Compared with GSPO~\citep{gspo} and GRPO~\citep{grpo}, SAPO provides both sequence-level coherence and token-level adaptivity: (1) Under mild assumptions—small on-policy steps and low dispersion of token log-ratios within a sequence—the average token gate concentrates to a smooth sequence-level gate $g(\log s_i(\theta))=\mathrm{sech}^2\!\big(\tfrac{\tau_i}{2}\log s_i(\theta)\big)$. Thus, SAPO reduces to a GSPO-like sequence formulation but with a continuous trust region. Crucially, when a few off-policy tokens push $s_i$ beyond GSPO’s hard band, GSPO suppresses gradients for the many near-on-policy tokens in that sequence, hurting sample efficiency.
SAPO, in contrast, preserves informative gradients by down-weighting only the offending tokens while keeping near-on-policy tokens influential. (2) Relative to GRPO, SAPO avoids hard token-level clipping that zeroes gradients outside a fixed range. Instead, SAPO scales updates smoothly, providing a more balanced way to retain useful learning signals while preventing unstable policy shifts. See more details in Section~\ref{sec:analysis}.

\paragraph{Why Different Temperatures for Positive and Negative Advantages}
\label{sec:pos_neg}

The hyperparameter $\tau$ controls the rate of attenuation: larger values produce faster decay. Although negative tokens are crucial for exploration and for preventing overfitting, they typically introduce greater instability than positive tokens. We justify this claim by analyzing how token-level gradients propagate through the logits. Let $z = [z_1, z_2, ..., z_{|\mathcal{V}|}]$ denote the logits (with vocabulary size $|\mathcal{V}|$), let $v$ denote a token, and compute output probabilities via a softmax operation, i.e., $\pi_\theta(v\mid q,y_{i,<t})=\exp(z_v)/\sum_{v'}\exp(z_{v'})$. We have
\begin{equation}
\begin{aligned}
\frac{\partial \log\pi_{\theta}(y_{i,t} \mid q, y_{i,<t})\,\widehat{A}_{i,t}}{\partial z_v}
& =
\frac{\partial \pi_{\theta}(y_{i,t} \mid q, y_{i,<t})}{\partial z_v} \cdot \frac{\widehat{A}_{i,t}}{\pi_{\theta}(y_{i,t} \mid q, y_{i,<t})} \\
& =
\frac{\mathbbm{1}(v = y_{i,t}) \exp(z_{y_{i,t}}) \sum_{v' \in \mathcal{V}} \exp(z_{v'}) - \exp(z_{y_{i,t}}) \exp(z_{v})}{\left( \sum_{v' \in \mathcal{V}} \exp(z_{v'})\right)^2} \cdot \frac{\widehat{A}_{i,t}}{\pi_{\theta}(y_{i,t} \mid q, y_{i,<t})}
\\
& =
\begin{cases}
 \bigl(1 - \pi_{\theta}(y_{i,t} \mid q, y_{i,<t})\bigr) \cdot \widehat{A}_{i,t} & \text{if } v = y_{i,t} \quad \text{(sampled token)} \\
-\pi_{\theta}(v \mid q, y_{i,<t}) \cdot \widehat{A}_{i,t} & \text{otherwise} \quad \text{(unsampled token)}
\end{cases}
\label{eq:pos_neg}
\end{aligned}
\end{equation}

Positive advantages increase the sampled token’s logit and decrease all unsampled logits; negative advantages do the opposite, raising the logits of many unsampled tokens. In RL fine-tuning for LLMs, the action space is a large vocabulary (often hundreds of thousands of tokens), whereas the number of desirable actions in a given state is small. Consequently, negative gradients diffuse to numerous irrelevant tokens—providing some regularization but also inducing instability, especially in off-policy scenarios. Accordingly, we use distinct temperatures for positive and negative tokens and set $\tau_{\text{neg}} > \tau_{\text{pos}}$ so that gradients on negative tokens decay more rapidly, thereby improving training stability and performance. 

\section{A Gating‑Function Perspective on SAPO’s Connections to GRPO and GSPO}
\label{sec:analysis}

\paragraph{Unified surrogate}
We consider a unified surrogate of the form
\begin{equation}
\mathcal{J}(\theta)
=
\mathbb{E}_{q\sim\mathcal{D},\,\{y_i\}_{i=1}^{G}\sim \pi_{\theta_\text{old}}(\cdot\mid q)}
\left[
\frac{1}{G}\sum_{i=1}^{G}
\frac{1}{|y_i|}\sum_{t=1}^{|y_i|}
f_{i,t}\big(r_{i,t}(\theta)\big)\;
\widehat{A}_{i,t}
\right],
\label{eq:unified_obj}
\end{equation}
where $f_{i,t}(\cdot)$ is an algorithm-specific gating function.
We further define the length-normalized sequence-level ratio as the geometric mean of token ratios:
\begin{equation}
\begin{aligned}
s_i(\theta)
=
\left(
\frac{\pi_\theta(y_i\mid q)}{\pi_{\theta_\text{old}}(y_i\mid q)}
\right)^{\!\frac{1}{|y_i|}}
=
\exp\!\left(
\frac{1}{|y_i|}\sum_{t=1}^{|y_i|}\log r_{i,t}(\theta)
\right),
\quad
s_{i,t}(\theta) 
= \mathrm{sg} \left[ s_{i}(\theta) \right]  \cdot \frac{ \pi_{\theta} (y_{i,t} | q, y_{i,<t}) }{ \mathrm{sg} \left[ \pi_{\theta} (y_{i,t} | q, y_{i,<t}) \right] },
\label{eq:seq_ratio}
\end{aligned}
\end{equation}
where $\mathrm{sg}[\cdot]$ denotes the stop gradient operation.

\paragraph{Algorithm-specific $f_{i,t}$}
The algorithms differ in the choice of $f_{i,t}$:
\begin{align}
\text{SAPO:}\quad
&f_{i,t}^{\mathrm{SAPO}}(r_{i,t}(\theta))
=
\frac{4}{\tau_i}\;\sigma\!\big(\tau_i\,(r_{i,t}(\theta)-1)\big),
\qquad
\tau_i=
\begin{cases}
\tau_{\mathrm{pos}}, & \widehat{A}_i>0,\\
\tau_{\mathrm{neg}}, & \widehat{A}_i\le 0,
\end{cases}
\label{eq:f_sapo}\\[0.75ex]
\text{GRPO:}\quad
&f_{i,t}^{\mathrm{GRPO}}(r_{i,t}(\theta);\widehat{A}_i)
=
\begin{cases}
\min\!\big(r_{i,t}(\theta),\;1+\varepsilon\big), & \widehat{A}_i>0,\\
\max\!\big(r_{i,t}(\theta),\;1-\varepsilon\big), & \widehat{A}_i\le 0,
\end{cases}
\label{eq:f_grpo}\\[0.75ex]
\text{GSPO:}\quad
&f_{i,t}^{\mathrm{GSPO}}(r_{i,t}(\theta);\widehat{A}_i)
\equiv
f_{i,t}^{\mathrm{seq}}\!\big(s_{i,t}(\theta);\widehat{A}_i\big)
=
\begin{cases}
\min\!\big(s_{i,t}(\theta),\;1+\varepsilon\big), & \widehat{A}_i>0,\\
\max\!\big(s_{i,t}(\theta),\;1-\varepsilon\big), & \widehat{A}_i\le 0.
\end{cases}
\label{eq:f_gspo}
\end{align}
Note that GSPO’s $f_{i,t}$ is token-invariant within a sequence, while SAPO and GRPO are token-dependent.

\paragraph{Gradient form for SAPO/GRPO}
Differentiating \eqref{eq:unified_obj} and using $\nabla_\theta r_{i,t}(\theta)=r_{i,t}(\theta)\nabla_\theta\log\pi_\theta(y_{i,t}\mid q,y_{i,<t})$, we obtain
\begin{equation}
\nabla_\theta \mathcal{J}(\theta)
=
\mathbb{E}\left[
\frac{1}{G}\sum_{i=1}^{G}
\frac{1}{|y_i|}\sum_{t=1}^{|y_i|}
f_{i,t}'\big(r_{i,t}(\theta)\big)\;
r_{i,t}(\theta)\;
\nabla_\theta\log\pi_\theta(y_{i,t}\mid q,y_{i,<t})\;
\widehat{A}_i
\right].
\label{eq:unified_grad}
\end{equation}

\subsection{SAPO–GSPO Connection: Reduction to a Sequence-Level Soft Gate}
\label{subsec:sapo_gspo}

We show that SAPO reduces to a GSPO-like sequence-level formulation under mild conditions, while retaining token-level adaptivity in heterogeneous sequences.

\paragraph{SAPO’s token-level soft gate}
Using $\sigma(x)(1-\sigma(x))=\frac{1}{(e^{x/2} + e^{-x/2})^2}=\tfrac{1}{4}\mathrm{sech}^2(x/2)$, we have
\begin{equation}
f_{i,t}^{\mathrm{SAPO}\,'}(r_{i,t}(\theta))
=
4\,\sigma\!\big(\tau_i\,(r_{i,t}(\theta)-1)\big)\Big(1-\sigma\!\big(\tau_i\,(r_{i,t}(\theta)-1)\big)\Big)
=
\mathrm{sech}^2\!\left(\frac{\tau_i}{2}\,(r_{i,t}(\theta)-1)\right),
\label{eq:sapo_gate_sech2}
\end{equation}

\paragraph{Assumptions}
We invoke two common assumptions:
\begin{enumerate}
\item[(A1)] Small-step/on-policy: $r_{i,t}(\theta)\approx 1$. Thus, $\log r_{i,t}(\theta)\approx r_{i,t}(\theta)-1$.
\item[(A2)] Low intra-sequence dispersion: letting $z_{i,t}(\theta):=\log r_{i,t}(\theta)$ and $\mu_i(\theta):=\tfrac{1}{|y_i|}\sum_t z_{i,t}(\theta)=\log s_i(\theta)$, the variance $\mathrm{Var}_i(\theta):=\tfrac{1}{|y_i|}\sum_t (z_{i,t}(\theta)-\mu_i(\theta))^2$ is small for most sequences.
\end{enumerate}

Under (A1), we have
\begin{equation}
f_{i,t}^{\mathrm{SAPO}\,'}(r_{i,t}(\theta))
=
\mathrm{sech}^2\!\left(\frac{\tau_i}{2}\,(r_{i,t}(\theta)-1)\right)
\;\approx\;
\mathrm{sech}^2\!\left(\frac{\tau_i}{2}\,\log r_{i,t}(\theta)\right)
=:g_{\tau_i}\!\big(z_{i,t}(\theta)\big).
\label{eq:gate_logr}
\end{equation}

\paragraph{Average token gates $\Rightarrow$ sequence gate}
By a second-order Taylor expansion of the smooth function $g_{\tau}(z)=\mathrm{sech}^2(\tfrac{\tau}{2}z)$ around $\mu_i(\theta)=\log s_i(\theta)$,
\begin{equation}
g_{\tau_i}(z_{i,t}(\theta))
=
g_{\tau_i}(\mu_i(\theta))
+
g'_{\tau_i}(\mu_i(\theta))\,(z_{i,t}(\theta)-\mu_i(\theta))
+
\frac{1}{2}\,g''_{\tau_i}(\xi_{i,t}(\theta))\,(z_{i,t}(\theta)-\mu_i(\theta))^2,
\label{eq:taylor_g}
\end{equation}
for some $\xi_{i,t}(\theta)$ between $z_{i,t}(\theta)$ and $\mu_i(\theta)$. Averaging over tokens cancels the linear term:
\begin{equation}
\frac{1}{|y_i|}\sum_{t=1}^{|y_i|} g_{\tau_i}(z_{i,t}(\theta))
=
g_{\tau_i}(\mu_i(\theta))
+
\frac{1}{2}\left(\frac{1}{|y_i|}\sum_{t=1}^{|y_i|} g''_{\tau_i}(\xi_{i,t}(\theta))\,(z_{i,t}(\theta)-\mu_i(\theta))^2\right).
\label{eq:avg_g}
\end{equation}
For $g_{\tau}(z)=\mathrm{sech}^2(\alpha z)$ with $\alpha=\tfrac{\tau}{2}$, a direct calculation gives
\begin{equation}
g''_{\tau}(z)
=
\alpha^2\Big(4\,\mathrm{sech}^2(\alpha z)-6\,\mathrm{sech}^4(\alpha z)\Big),
\quad
\sup_{z}|g''_{\tau}(z)|=2\alpha^2=\frac{\tau^2}{2}.
\label{eq:g2_bound}
\end{equation}
Hence, the average token gate is well-approximated by the sequence gate with a uniform bound:
\begin{equation}
D_{i}(\theta)=
\left|
\frac{1}{|y_i|}\sum_t g_{\tau_i}(z_{i,t}(\theta)) - g_{\tau_i}(\mu_i(\theta))
\right|
\le\;
\frac{1}{2}\,\sup_{z}|g_{\tau_i}''(z)|\;
\mathrm{Var}_i(\theta)
=
\frac{\tau_i^2}{4}\;\mathrm{Var}_i(\theta).
\label{eq:avg_gate_bound}
\end{equation}

Starting from \eqref{eq:unified_grad} and applying $r_{i,t}(\theta)\approx 1$ (A1), we have
\begin{equation}
\nabla_\theta \mathcal{J}_{\mathrm{SAPO}}
\approx
\mathbb{E}\left[
\frac{1}{G}\sum_{i=1}^{G}
\frac{1}{|y_i|}\sum_{t=1}^{|y_i|}
g_{\tau_i}(z_{i,t}(\theta))\;
\nabla_\theta\log\pi_\theta(y_{i,t}\mid q,y_{i,<t})\;\widehat{A}_i
\right].
\label{eq:sapo_grad_step1}
\end{equation}
Using \eqref{eq:avg_gate_bound}, we have
\begin{equation}
\begin{aligned}
\nabla_\theta \mathcal{J}_{\mathrm{SAPO}}
& \approx
\mathbb{E}\left[
\frac{1}{G}\sum_{i=1}^{G}
g_{\tau_i}\!\big(\log s_i(\theta)\big)\;
\left(\frac{1}{|y_i|}\sum_{t=1}^{|y_i|}\nabla_\theta\log\pi_\theta(y_{i,t}\mid q,y_{i,<t})\right)\;
\widehat{A}_i
\right] \\
& =
\mathbb{E}\left[
\frac{1}{G}\sum_{i=1}^{G}
g_{\tau_i}\!\big(\log s_i(\theta)\big)\;
\nabla_\theta\log s_i(\theta)\;\widehat{A}_i
\right].
\label{eq:sapo_seq_form}
\end{aligned}
\end{equation}
Thus, under (A1)–(A2), SAPO reduces to a sequence-level update structurally similar to GSPO with a smooth gate $g_{\tau_i}(\log s_i(\theta))=\mathrm{sech}^2(\tfrac{\tau_i}{2}\log s_i(\theta))$.

\paragraph{Do the two assumptions (A1) and (A2) hold?} 

\begin{figure}[thb] 
    \centering 
    \begin{minipage}{0.33\textwidth}
        \centering
        \includegraphics[width=\linewidth]{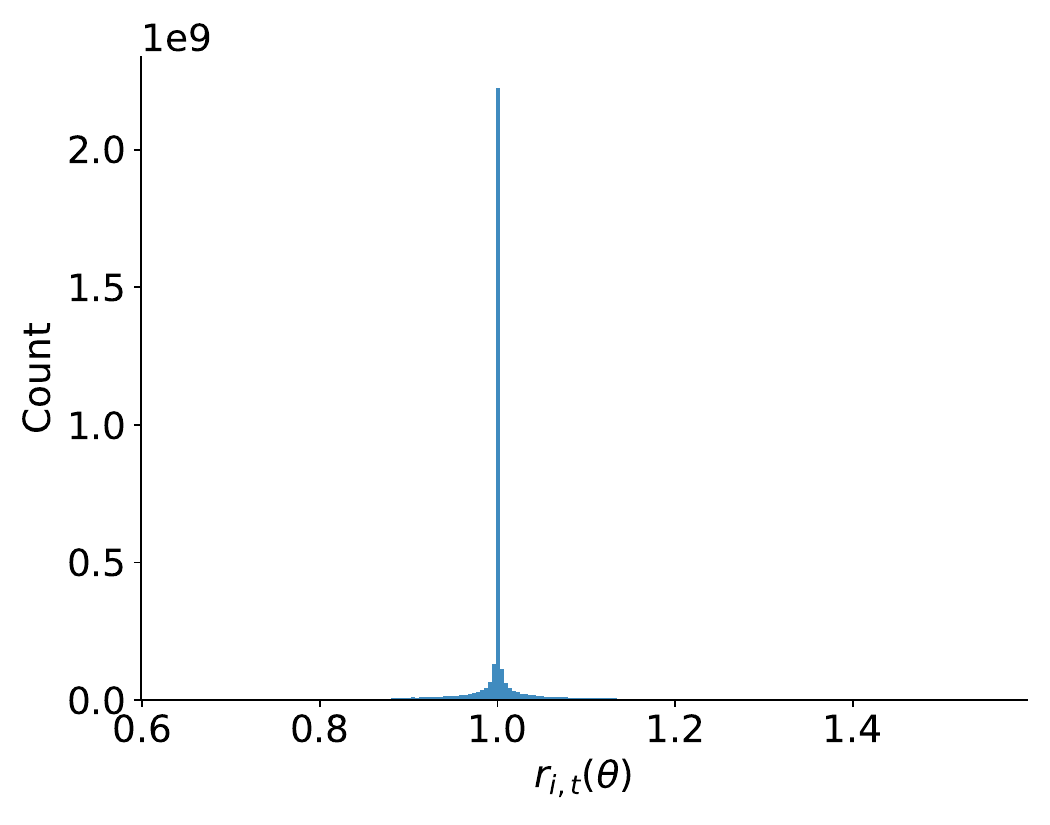}
    \end{minipage}%
    \hfill 
    \begin{minipage}{0.33\textwidth}
        \centering
        \includegraphics[width=\linewidth]{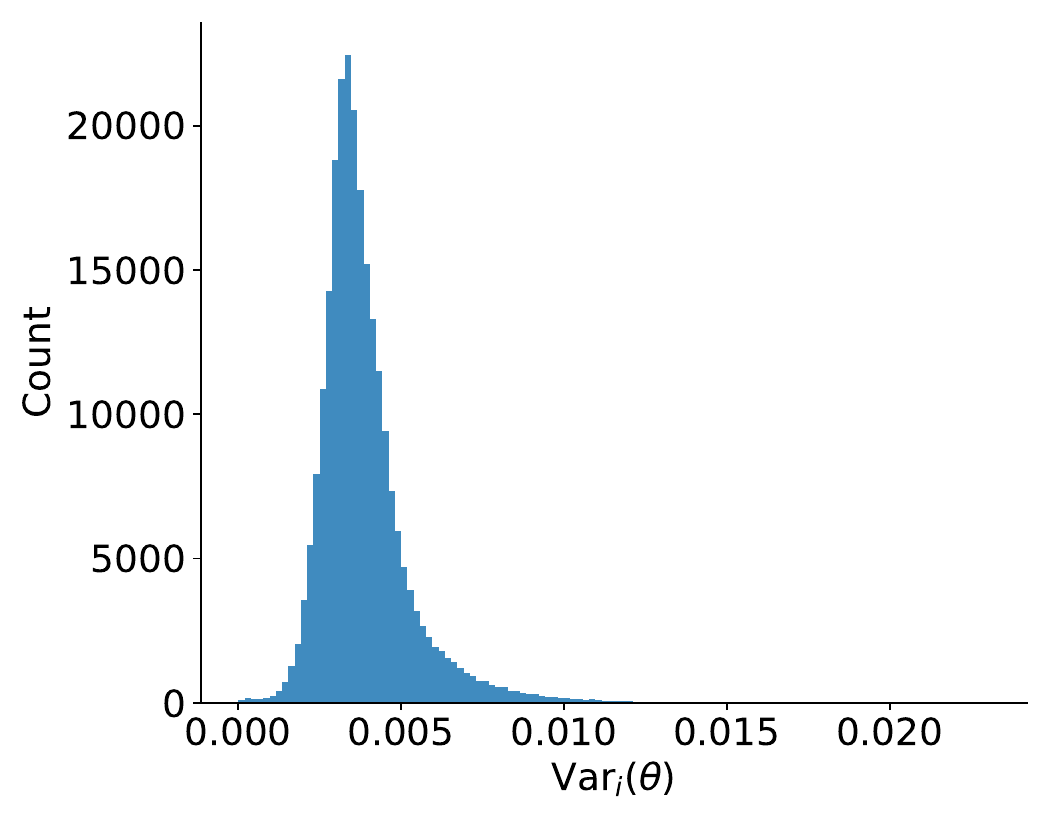}
    \end{minipage}%
    \hfill 
    \begin{minipage}{0.33\textwidth}
        \centering
        \includegraphics[width=\linewidth]{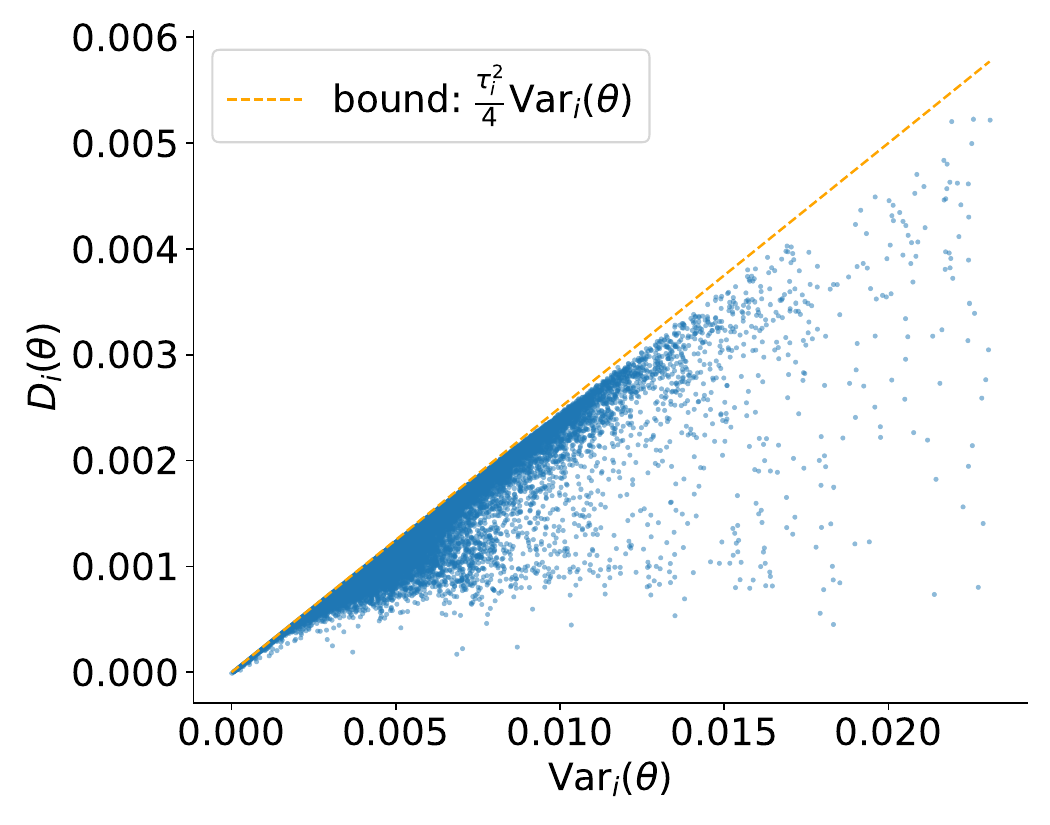}
    \end{minipage}
    \caption{Empirical validation of assumptions (A1)–(A2) on the MoE model (Qwen3-30B-A3B). Left: histogram of token importance ratios $r_{i,t}(\theta)$. Middle: histogram of per-sequence log-ratio variance $\mathrm{Var}_i(\theta)$. Right: scatter of $\mathrm{Var}_i(\theta)$ versus $D_i(\theta)$.}
    \label{fig:MOE}
\end{figure}

\begin{figure}[thb] 
    \centering 
    \begin{minipage}{0.33\textwidth}
        \centering
        \includegraphics[width=\linewidth]{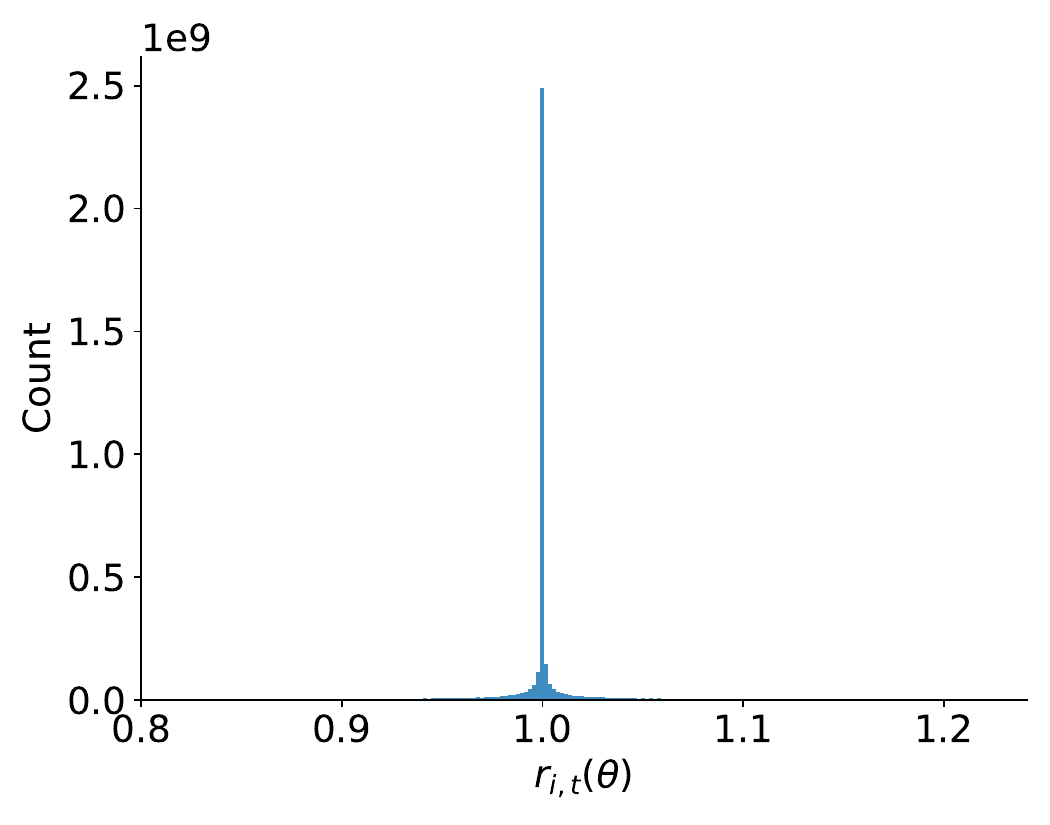}
    \end{minipage}%
    \hfill 
    \begin{minipage}{0.33\textwidth}
        \centering
        \includegraphics[width=\linewidth]{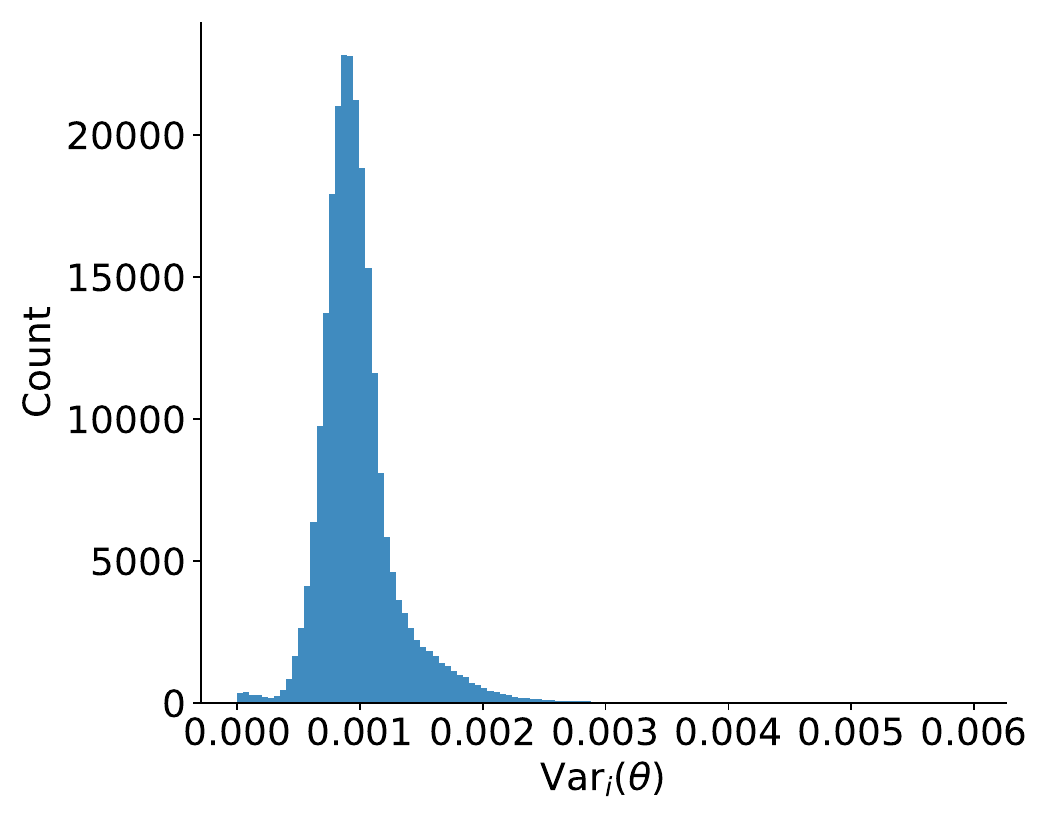}
    \end{minipage}%
    \hfill 
    \begin{minipage}{0.34\textwidth}
        \centering
        \includegraphics[width=\linewidth]{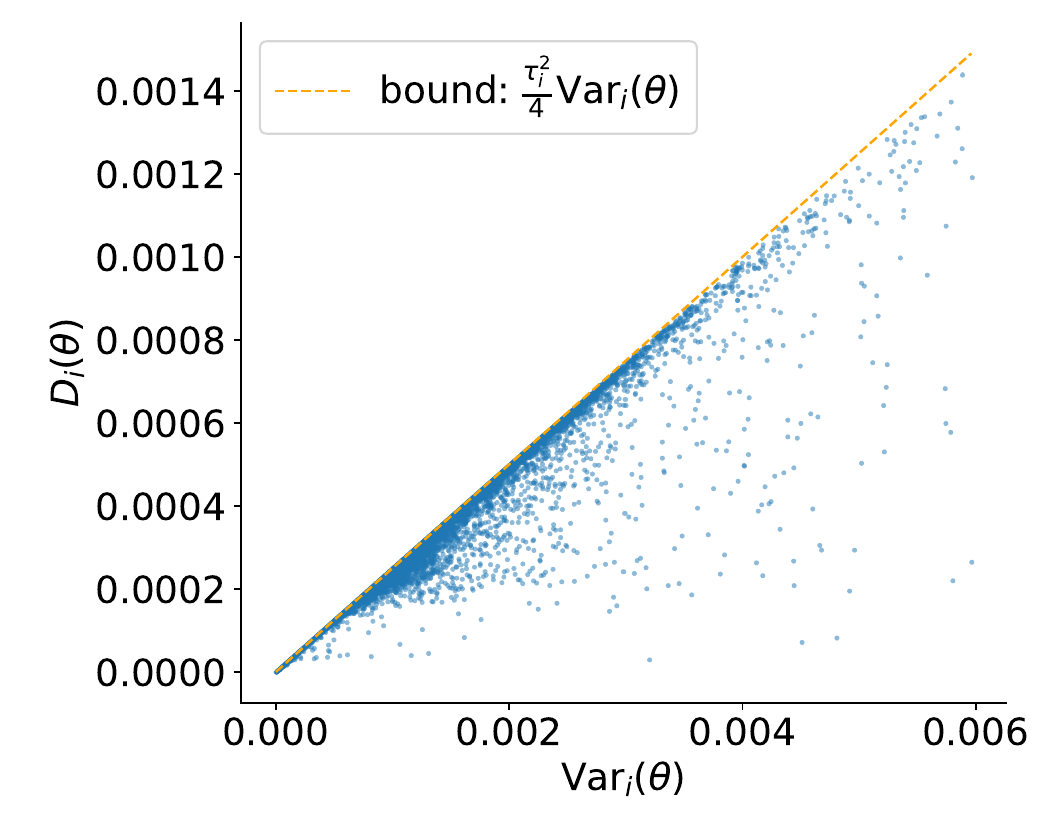}
    \end{minipage}
    \caption{Empirical validation of assumptions (A1)–(A2) on the dense model (Qwen3-4B). Left: histogram of token importance ratios $r_{i,t}(\theta)$. Middle: histogram of per-sequence log-ratio variance $\mathrm{Var}_i(\theta)$. Right: scatter of $\mathrm{Var}_i(\theta)$ versus $D_i(\theta)$.}
    \label{fig:Dense}
\end{figure}

We empirically assess the small-step assumption (A1) and the low intra-sequence dispersion assumption (A2) by plotting the histograms of token ratios $r_{i,t}(\theta)$ and per-sequence log-ratio variance $\mathrm{Var}_i(\theta)$ in Figures \ref{fig:MOE} and \ref{fig:Dense} for both MoE and dense models. The MoE model is a cold-start checkpoint of Qwen3-30B-A3B, and the dense model is a cold-start checkpoint of Qwen3-4B. The statistics are computed over more than $10^5$ sequences and $10^9$ tokens drawn from off-policy mini-batches. We observe that $r_{i,t}(\theta)$ is sharply concentrated around $1$, and $\mathrm{Var}_i(\theta)$ typically remains below $0.02$, with a relatively wider distribution for the MoE model (likely reflecting heterogeneity induced by expert routing) and a tighter concentration for the dense model. These distributions indicate that (A1) and (A2) hold in the majority of cases, especially for dense architectures. Moreover, small $D_{i}(\theta)$ directly implies that the average token gate is well-approximated by the sequence-level gate, supporting our reduction.

\paragraph{Advantages over GSPO} Compared with GSPO, SAPO has the following advantages: (1) \emph{Smoothness and stability.} The soft gate varies continuously with the sequence deviation, avoiding the discontinuities of hard clipping and reducing optimization noise. (2) \emph{Token-level adaptivity with sequence-level coherence.} Under (A1) and (A2), SAPO acts like a sequence-level method; when these conditions are violated (heterogeneous tokens or outliers), SAPO defaults to its token-level gating, selectively down-weighting outliers while preserving informative tokens—an ability GSPO lacks.

\subsection{SAPO–GRPO Connection: Smooth Token Gates vs. Hard Token Clipping}
\label{subsec:sapo_grpo}

\paragraph{GRPO’s piecewise-hard token gate}
For GRPO, $f_{i,t}^{\mathrm{GRPO}}(r_{i,t}(\theta);\widehat{A}_i)$ is piecewise constant with respect to the clipping band. Differentiating yields
\begin{equation}
f_{i,t}^{\mathrm{GRPO}\,'}(r_{i,t}(\theta);\widehat{A}_i)
=
\begin{cases}
1, & \widehat{A}_i>0 \text{ and } r_{i,t}(\theta) \le 1+\varepsilon,\\
0, & \widehat{A}_i>0 \text{ and } r_{i,t}(\theta) > 1+\varepsilon,\\
1, & \widehat{A}_i\le 0 \text{ and } r_{i,t}(\theta) \ge 1-\varepsilon,\\
0, & \widehat{A}_i\le 0 \text{ and } r_{i,t}(\theta) < 1-\varepsilon.
\end{cases}
\label{eq:grpo_gate_deriv}
\end{equation}

Hence, GRPO employs a binary trust region: tokens inside receive the same gradient as that of the unclipped objective; tokens outside receive zero gradient.

\paragraph{Advantages over GRPO}
Compared with GRPO, SAPO replaces the hard indicator in \eqref{eq:grpo_gate_deriv} with the smooth kernel $f_{i,t}^{\mathrm{SAPO}\,'}(r_{i,t}(\theta))=\mathrm{sech}^2(\tfrac{\tau_i}{2}(r_{i,t}(\theta)-1))$, which avoids gradient vanishing and enables more stable update dynamics. When the policy change is small, gradients remain responsive and permit larger parameter updates; as the deviation grows, gradients shrink smoothly, resulting in more conservative adjustments. In contrast, GRPO’s hard token clipping yields an all‑or‑nothing gate, often leading to brittle and unstable optimization behavior.

\subsection{Summary}

These RL algorithms primarily differ in how they handle off-policy tokens for which $r_{i,t}(\theta)$ deviates from $1$. From a token-level perspective, SAPO provides a smooth downweighting mechanism; from a sequence-level perspective, SAPO suppresses gradients from extreme off-policy tokens in sequences, thereby constructing more effective sequences for training. In contrast, GRPO and GSPO rely on hard clipping, which is less adaptive than SAPO for optimization.

\begin{figure}[tbp]
    \centering
    \includegraphics[width=0.9\linewidth]{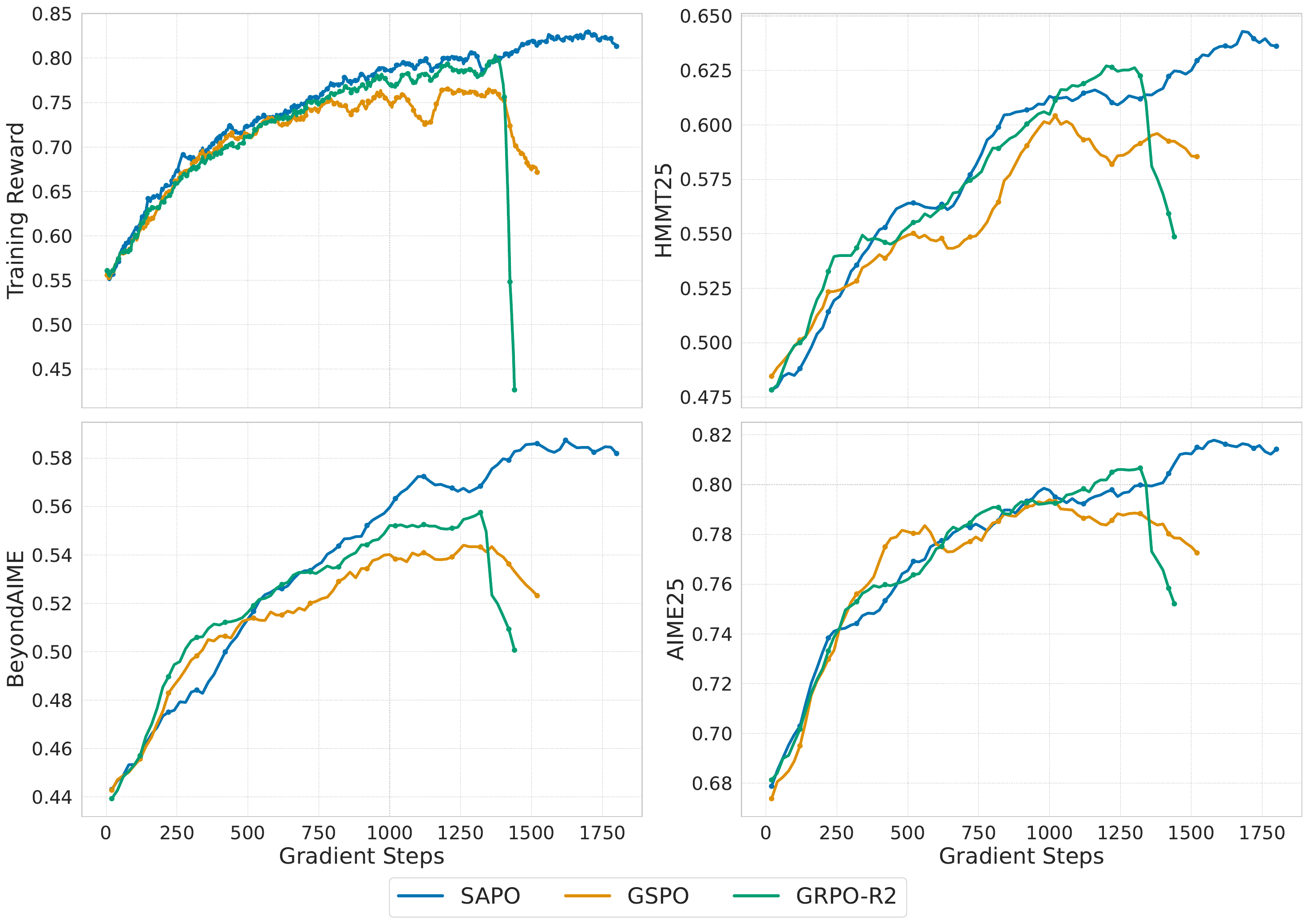}
    \caption{
Training reward and validation performance of a cold-start model fine-tuned from Qwen3-30B-A3B-Base under different RL algorithms. SAPO exhibits consistently stable learning and achieves higher final performance compared with GSPO and GRPO-R2, both of which experience early training collapse.
}
    \label{fig:results_algo}
\end{figure}

\section{Experiments}
\subsection{Controlled Experiments}
We conduct experiments using a cold-start model fine-tuned from Qwen3-30B-A3B-Base on mathematical reasoning queries. We report both the training reward and the validation performance (average Pass@1 over 16 samples) on the AIME25 \citep{aime25}, HMMT25 \citep{hmmt25}, and BeyondAIME \citep{beyondaime} benchmarks. During RL training, each batch of rollout data is divided into four mini-batches for gradient updates.
For SAPO, we set $\tau_{\text{pos}} = 1.0$ and $\tau_{\text{neg}} = 1.05$ in Equation~\eqref{eq:tau}. We compare SAPO with GSPO and GRPO-R2 (i.e., GRPO equipped with routing replay), using the same hyperparameter configurations as in \citet{gspo}.

Figure~\ref{fig:results_algo} shows that SAPO consistently improves model performance across all benchmarks, achieving both higher stability and stronger final performance compared with GSPO and GRPO-R2. While GSPO and GRPO-R2 exhibit early-stage training collapse, SAPO maintains stable training dynamics and ultimately attains superior performance. Notably, SAPO does not rely on routing replay for stabilization or strong performance, which improves exploration and reduces engineering overhead for RL systems.

\begin{figure}[thbp]
    \centering
    \includegraphics[width=0.9\linewidth]{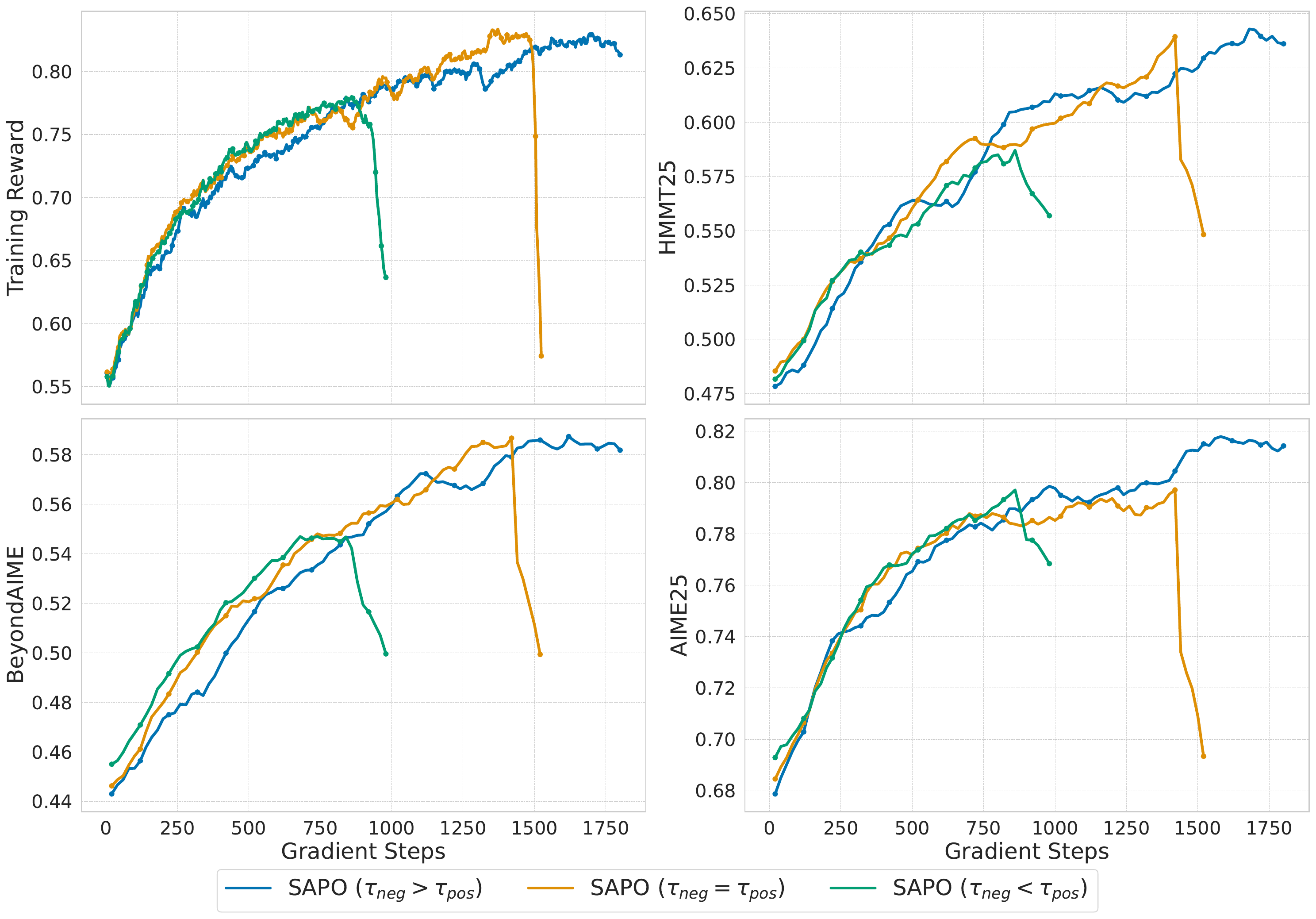}
    \caption{
Training reward and validation performance of a cold-start model fine-tuned from Qwen3-30B-A3B-Base using SAPO with different temperature settings. Using a higher temperature for negative tokens ($\tau_{\text{neg}} > \tau_{\text{pos}}$) leads to the most stable training dynamics, whereas setting $\tau_{\text{neg}} < \tau_{\text{pos}}$ causes significant instability.
}
    \label{fig:results_sapo}
\end{figure}

To empirically examine the effect of choosing $\tau_{\text{neg}} > \tau_{\text{pos}}$, we evaluate three configurations: $\tau_{\text{neg}} = 1.05 > \tau_{\text{pos}} = 1.0$, $\tau_{\text{neg}} = \tau_{\text{pos}} = 1.0$, and $\tau_{\text{neg}} = 0.95 < \tau_{\text{pos}} = 1.0$. As shown in Figure~\ref{fig:results_sapo}, training is most stable when negative tokens are assigned a higher temperature ($\tau_{\text{neg}} = 1.05$) and most unstable when they are assigned a lower temperature ($\tau_{\text{neg}} = 0.95$). These results suggest that gradients associated with negative tokens contribute more strongly to training instability, and that SAPO's asymmetric temperature design effectively mitigates this issue.

\subsection{Qwen3‑VL Training}

We apply SAPO to train models in the Qwen3-VL family to assess its effectiveness in practical large-scale settings. Our experiments show that SAPO consistently improves performance across models of varying sizes and across both MoE and dense architectures. We train on a broad collection of text and multimodal tasks, including mathematics, coding, and logical reasoning. To support multi-task learning, we maintain a fixed sampling ratio for each task within each batch. We also employ a large batch size, splitting each batch of rollout data into two mini-batches for gradient updates, ensuring that each mini-batch provides sufficient learning signal for all tasks.

\begin{figure}[htbp]
    \centering
    \includegraphics[width=0.9\linewidth]{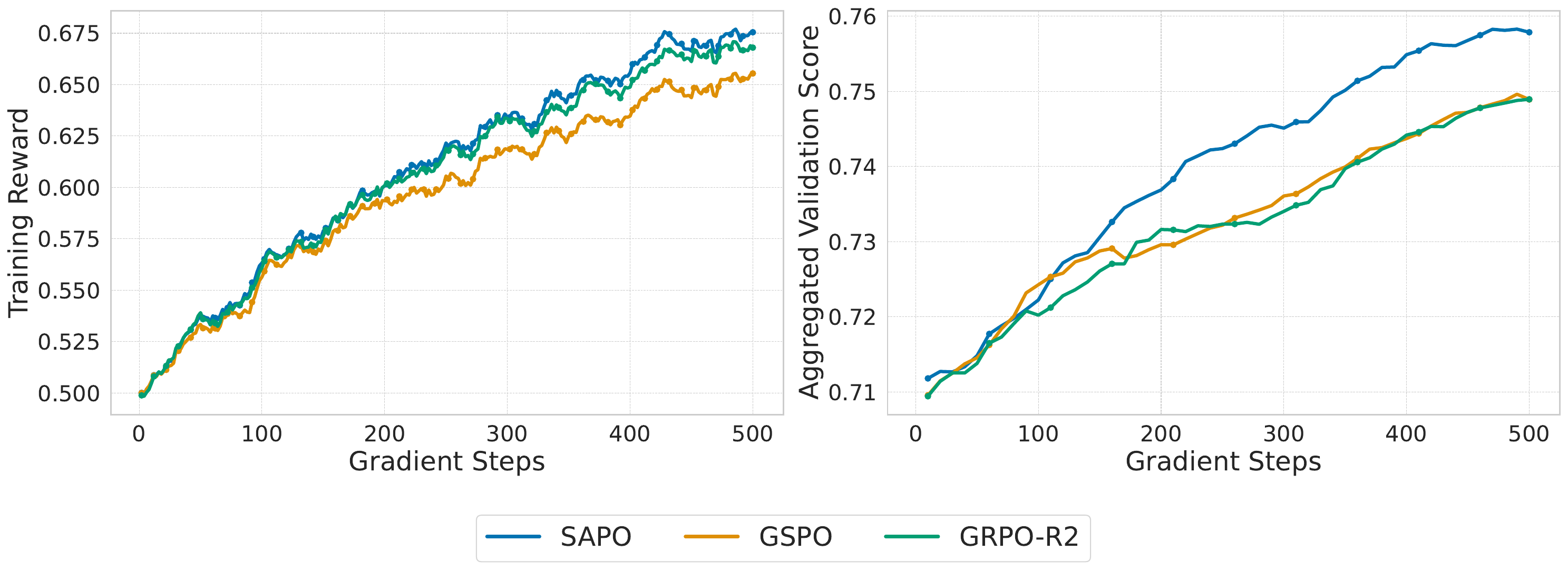}
    \caption{
Training reward and validation performance of Qwen3-VL-30B-A3B from a preliminary cold-start initialization, showing that SAPO achieves consistent improvements and outperforms GSPO and GRPO‑R2 under the same compute budget.
}
    \label{fig:results_vl}
\end{figure}

To highlight SAPO’s advantages over GSPO and GRPO-R2, we evaluate all three reinforcement learning algorithms starting from a preliminary cold-start checkpoint of Qwen3-VL-30B-A3B. We report training rewards and mean validation performance on four benchmarks: AIME25~\citep{aime25} (Pass@1 with 32 samples), LiveCodeBench v6~\citep{jain2024livecodebench} (Pass@1 with 8 samples), ZebraLogic~\citep{lin2025zebralogic}, and MathVision~\citep{wang2024measuring}. As shown in Figure~\ref{fig:results_vl}, SAPO achieves steady performance gains throughout training and outperforms both baselines under equal compute budgets.

\section{Conclusion}

We introduce Soft Adaptive Policy Optimization (SAPO), a smooth and token-adaptive reinforcement learning algorithm designed to address the instability and inefficiencies associated with hard-clipped policy optimization in large language models. By replacing discontinuous clipping with a temperature-controlled soft gate and employing asymmetric temperatures to better regulate negative-token gradients, SAPO provides a more stable and informative optimization signal. Empirical results on several mathematical reasoning benchmarks demonstrate that SAPO extends the duration of stable training and achieves higher Pass@1 performance under comparable budgets. Beyond controlled settings, large-scale experiments on Qwen3‑VL models further demonstrate that SAPO delivers consistent improvements across diverse text and multimodal tasks and across different model sizes and architectures. These findings suggest that smooth and adaptive gating mechanisms offer a promising direction for improving the robustness and effectiveness of RL training for large language models.

\end{document}